\documentclass[twocolumn]{article}
\usepackage[margin=0.75in]{geometry}
\usepackage{graphicx}
\usepackage{float}
\usepackage{xcolor}
\usepackage{geometry}
\usepackage{multirow}
\usepackage{url}
\usepackage{xurl} %fixed overflow urls
\usepackage{bbold}
\usepackage{mathtools}
\usepackage{natbib}
\usepackage{hyperref}

\usepackage{amsmath}
\usepackage{algorithm}
\usepackage{algpseudocode}
\usepackage{xcolor}         % for the \note{} command
\usepackage{titlesec}
\titleformat*{\section}{\large\bfseries}
\titleformat*{\subsection}{\large\bfseries}
\titleformat*{\subsubsection}{\large\bfseries}

\algnewcommand\algorithmicforeach{\textbf{for each}}
\algdef{S}[FOR]{ForEach}[1]{\algorithmicforeach\ #1\ \algorithmicdo}
\newtheorem{ass}{Assumption}

\hypersetup{
    pdftitle={Efficient Heterogeneous Treatment Effect Estimation With Multiple Experiments and Multiple Outcomes},
    pdfauthor={Yao et. al},
}

\begin{document}
\title{Efficient Heterogeneous Treatment Effect Estimation With Multiple Experiments and Multiple Outcomes}
\author{
    Leon Yao\thanks{Massachusetts Institute of Technology}
    \ \ \ 
    Caroline Lo\thanks{Facebook}
    \ \ \ 
    Israel Nir\footnotemark[2]
    \ \ \ 
    Sarah Tan\footnotemark[2]
    \and
    Ariel Evnine\footnotemark[2]
    \ \ \ 
    Adam Lerer\footnotemark[2]
    \ \ \ 
    Alex Peysakhovich\footnotemark[2]
}
\maketitle
% !TEX root = main.tex

\begin{abstract}
Learning heterogeneous treatment effects (HTEs) is an important problem across many fields. Most existing methods consider the setting with a single treatment arm and a single outcome metric. However, in many real world domains, experiments are run consistently - for example, in internet companies, A/B tests are run every day to measure the impacts of potential changes across many different metrics of interest. We show that even if an analyst cares only about the HTEs in one experiment for one metric, precision can be improved greatly by analyzing all of the data together to take advantage of cross-experiment and cross-outcome metric correlations. We formalize this idea in a tensor factorization framework and propose a simple and scalable model which we refer to as the low rank or LR-learner. Experiments in both synthetic and real data suggest that the LR-learner can be much more precise than independent HTE estimation.
\end{abstract}
% !TEX root = main.tex

\section{Introduction}
Constructing personalized policies is increasingly of interest across many diverse communities ranging from internet companies to personalized medicine \citep{peysakhovich2016combining, kent2018personalized}. Unfortunately, estimating heterogeneous effects, sometimes referred to as conditional average treatment effects (CATEs), requires much more statistical power than simply estimating the average treatment effect \citep{gelman_2018}. In this paper, we introduce the LR-learner, a method of combining multiple experiments and multiple outcome metrics to improve the precision of CATE estimation across all of them. Even if an analyst is only interested in estimating CATEs for a single experiment, adding data from past experiments allows a more efficient estimation. 

At its core, the goal of heterogeneous estimation is to output an estimate of the treatment effect for a unit whose features are specified by the  vector $x$. Under the potential outcomes framework \citep{rubin1974estimating}, we denote the potential outcomes under treatment and control as $Y_1$ and $Y_0$, and their conditional means as $\mu_1(x) = E[Y_1|X=x]$ and $\mu_0(x) = E[Y_0|X=x]$, respectively. Our estimand of interest is the $$\text{CATE}(x) = \tau(x) = \mu_1(x) - \mu_0(x).$$ 

Much of the literature on experimental analysis focuses on the case of a single experiment with a single treatment and a single outcome (e.g. a medical trial of a drug with survival as an outcome). Here there is a large body of work on efficiently estimating both the average effect of this experiment as well as finding subgroups which are more or less affected by the treatment \citep{imbens2015causal, wager2018estimation,kunzel2019metalearners}.

In many environments of interest, however, we are not considering a single experiment/outcome in  isolation. For example, the A/B testing that is common in modern internet companies is different than the classical case in a few ways. First, large companies can run hundreds of tests per day \citep{bezos, xie2016improving}, measuring and trading off multiple outcome measures. For example, a movie streaming company can conduct A/B tests that run the gamut from tweaking movie recommendation algorithm parameters, to testing landing page layout changes. Each of these experiments can be evaluated on a set of metrics ranging from user retention to server CPU usage. Decision makers then trade off between metrics to make a deployment or non-deployment decision. The effect of each intervention is likely to be heterogeneous across treated units. Second, unlike in many classical applications, the set of unit level features that analysts have access to is very high dimensional. Thus a key problem is one of representation learning: finding a good, lower dimensional, feature description for efficiently learning the heterogeneity of treatment effects. 

Our key insight is that pooling information across multiple experiments and multiple outcomes and learning a single feature representation for treated units allows us to greatly improve the precision with which we estimate heterogeneous effects over standard methodologies. The potential to pool experiments is increasingly viable in the online A/B testing setting. In addition, the idea of multiple experiments with high dimensional feature spaces is becoming applicable to other domains e.g. pooling many clinical trials or experimental work in systems biology.

For the one experiment/one outcome case, a simple methodology for learning CATEs is the T-learner \citep{kunzel2019metalearners}, where we train a model, $\hat{\mu}_1(x)$, on data from units in the treatment group and train a model, $\hat{\mu}_0(x)$, on data from units in the control group. The estimated effect can be computed as $\hat{\tau}(x) = \hat{\mu}_1(x) - \hat{\mu}_0(x)$.

Here, we build on the T-learner and consider the case of multiple experiments and multiple outcomes. Every unit may be subject to multiple experiments, and all outcome metrics are observed. Our goal is to estimate the treatment effect for all units across all experiments. This task can be modeled as a tensor factorization problem and can be done efficiently when the tensor is low rank. We refer to our method as the Low Rank Learner, or LR-learner.

For each unit, the LR-learner learns a $d$-dimensional unit embedding using a small neural network over input unit level features. The LR-learner also learns an embedding for each experiment treatment and control. Finally, for each outcome, a linear operator transforms the experiment and unit embeddings so that the dot product of these transformed embeddings predicts the desired outcome. The neural network, experiment embeddings and operators are learned jointly using the observed data. 

The LR-learner takes advantage of the low dimensional bottleneck of the embeddings and linear operators, which is what allows us to pool information across experiments and outcome types. In practice, true CATEs are not observed (since we do not observe any unit in both treatment and control), so we first evaluate the LR-learner on fully synthetic and semi-synthetic data. We show that pooling information across experiments and outcomes greatly improves efficiency relative to independent learning.

Finally, we evaluate the LR-learner on a real dataset of $53$ product experiments run at a large internet company, focusing on $5$ real key performance indicators used by teams to make product decisions. We show that there is substantial information in heterogeneous effects shared across experiments and metrics, and that the LR-learner outperforms the independent level T-learner. In addition, we discuss other real world advantages of the LR-learner including scalability and the ability to use the LR-learner as a feature extractor negating the need for constant retraining when new experiments are run.
% !TEX root = main.tex
\section{Related Work}
There exist many methods for estimating CATEs on data from a single experiment \citep{peysakhovich2016combining,powers2018some,wager2018estimation,kunzel2019metalearners,kennedy2020optimal}. Our main contribution over this large literature is the consideration of combining data from multiple experiments and multiple metrics into a single model. Although there are many methods that aim to directly estimate the treatment effect \citep{athey2018generalized}, in this paper we consider only the Two-Model approach \citep{pmlr-v67-gutierrez17a}, because it naturally extends to the multiple experiment case. Possible extensions of our ideas to other ``meta-learners,'' like the X-learner \citep{kunzel2019metalearners} are an interesting future direction though beyond the scope of this paper.

There is also a lot of recent literature on matrix completion methods on panel data \citep{agarwal2021causal, agarwal2021synthetic, athey_causal_factorization, bai2021matrix, fernandezval2021lowrank}. They consider the setting in which units are observed over multiple time periods. In each period, a unit can be exposed to some binary treatment, and their outcome for that period is recorded. These papers assume some sort of sparsity structure for when units receive treatment or when outcomes are observed. Our methodology does not assume any sparsity structure, so is not reliant on panel data, but can also be applied to this setting.

The model architecture employed by the LR-learner is inspired by the RESCAL model used for tensor factorization \citep{lerer2019pytorch, nickel2011three}. Typically, these methods are applied to construct low dimensional embeddings of objects such as social or knowledge graphs \citep{lerer2019pytorch}. So, a conceptual contribution of our paper is to make a link between two concepts. First, we frame the causal inference problem as a missing data problem \citep{ding2018causal}. Second, to solve the missing data problem, we use efficient low dimensional methods to fill in unobserved tensor values. We note that unlike in the case of a matrix, where the singular value decomposition provides the ``canonical'' way to factorize, there are many ways to factorize a tensor \citep{welling2001positive}. Exploring other tensor factorization methods is an interesting avenue for future work.

% !TEX root = main.tex

\section{Potential Outcomes as Tensor Factorization}

We consider treatment assigned at the level of an individual (aka. unit), and consider the presence of multiple outcome metrics. Let $n$ be the number of units, $m$ be the number of observed pre-treatment features, $K$ be the number of experiments, and $J$ be the number of observed outcome metrics. We denote by $Y_{ijkt}$ the potential outcome of unit $i$ on metric $j$ when assigned to experiment $k$ with treatment arm $t$. We denote $t=0$ to be the `control' arm for any experiment.

We make the following assumptions:

\begin{ass}[SUTVA \citep{rubin1978bayesian}]
Every observed outcome for unit $i$ is independent of the actual treatment assignment of other units $j \neq i$.
\end{ass}

\begin{ass}[Random Assignment]
In any data we observe, units are assigned to treatments (or control) unconditional on any observed or unobserved feature values.
\end{ass}

The individual treatment effect (ITE) for individual $i$ for metric $j$ in experiment $k$ for treatment $t$ is denoted
$$\text{ITE}_{ijk} = Y_{ijkt} -  Y_{ijk0}.$$

In this framework, a typical `classical' randomized trial (e.g. a medical trial) would have a single arm, a single outcome, and a feature vector for each unit. In this case, the ITE would simply be denoted by a vector with one value (i.e treatment effect estimate) per unit $i$. 

If we move to the case of a standard trial with multiple outcome measures the ITE can be represented as a matrix with rows corresponding to individuals and columns corresponding to outcome measures. Finally, moving to the fully general multi-experiment, multi-treatment, multi-outcome case the ITE becomes a tensor. This brings us to our final key assumption:

\begin{ass}
The ITE tensor has a low rank structure.
\end{ass}

Assuming the ITE tensor has low rank structure means that there is information about a given entry of interest in other entries. This is the core of the proposed methodology. In practice, we do not observe any ITEs since no individual is ever in both a treatment and a control simultaneously. Instead, we will focus on a relative of the ITE which instead featurizes each unit, $i$, using a vector, $x_i$. The CATE for treatment arm, $t$, in experiment, $k$, is defined as $$\text{CATE}_{ijkt}(X_i) = E[Y_{ijkt} \mid X_i] - E[Y_{ijk0} \mid X_i].$$

\noindent If the ITE tensor is low rank, then observing CATEs for experiment $k$ and metric $j$ tells us something about CATEs in other experiments/metrics/treatments. This is precisely what the LR-learner will take advantage of.

\section{LR-Learner for Heterogeneous Effects}
We now discuss how to train the LR-learner as a tensor factorization. Our factorization has several components: We have a sample of units assigned to experiments/treatments and their features denoted by $\mathcal{D} = \lbrace y_{ijkt}, x_i \rbrace$. We train a featurized tensor factorization model which includes 

\begin{enumerate}
    \item A $d$ dimensional representation of each unit $i$, which is a function of $x_i$, called $v (x_i) \in R^{d}$. This is expressed as a simple neural network which inputs $x_i$ and outputs a $d$ dimensional feature vector $v_i$.
    \item A $d$ dimensional representation of each experiment treatment arm $t$ and control $e^t_k \in R^{d}$ and $e^0_k \in R^{d}$.
    \item A $d \times d$ transformation matrix for each outcome $A_j$.
    \item The parameters are trained with an objective similar to RESCAL \citep{nickel2011three}     $$\min_{v, A, e} \sum_{(y_{ijkt},x_i) \in \mathcal{D}} (y - v (x_i)' A_{j} e^t_{k})^2.$$
\end{enumerate}

\begin{figure}[]
\centering
\includegraphics[width=0.49\textwidth]{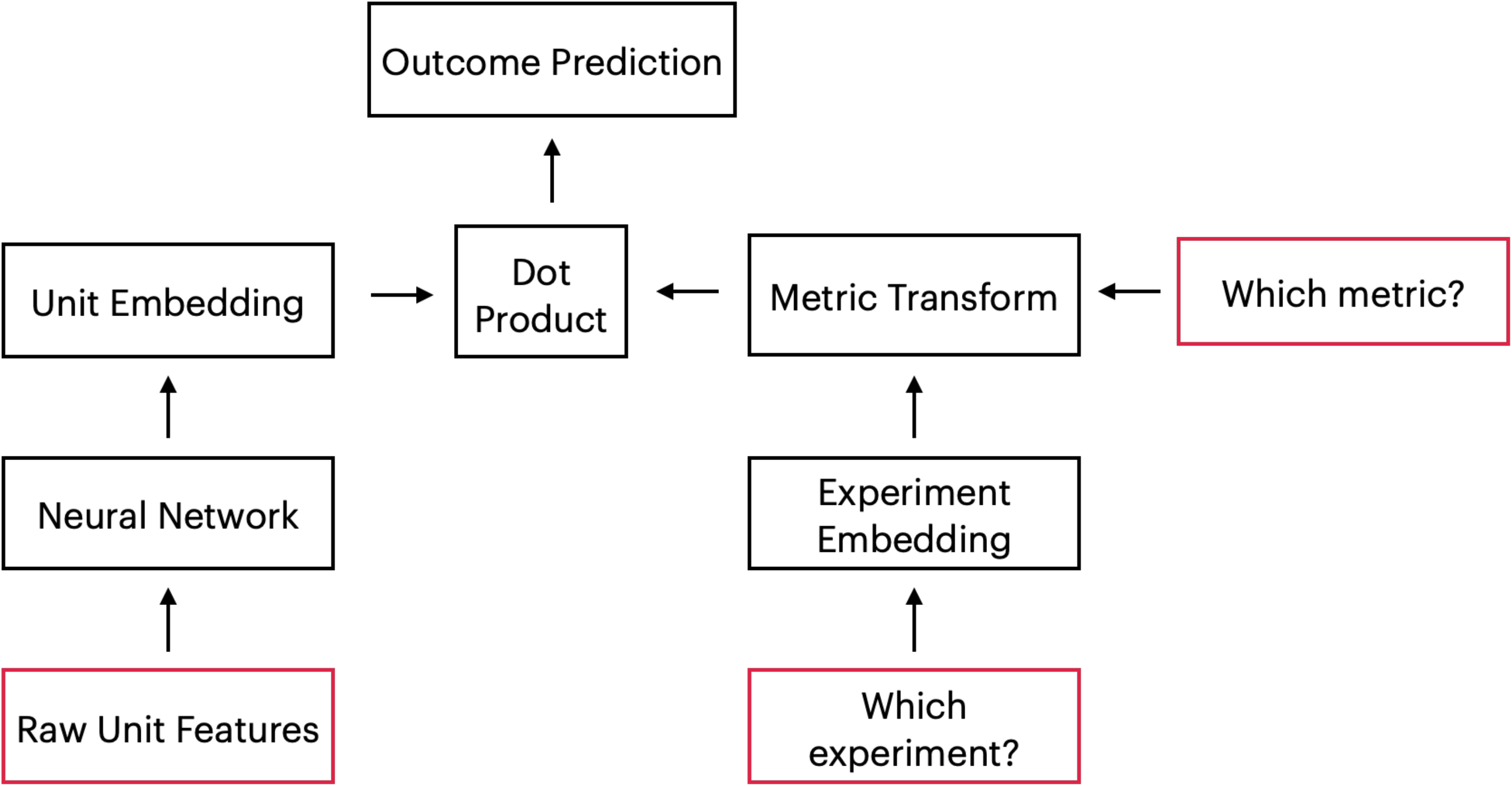}
\caption{Visual description of the LR-learner model. Red boxes indicate inputs transformed to give an output of the form $y_{ijkt} = v(x_i)' A^k e^t_k$.}\label{nn_fig}
\end{figure}

\noindent The LR-learner structure is graphically summarized in the figure \ref{nn_fig}.

Given the trained model, the predicted potential outcome for a unit $i$ in experiment $j$ in arm $t$ for outcome $k$ is given by:
$$\hat{\mu}^t_{jk}(x_i) = v(x_i)' A_j e^t_{k}$$ 

This means the estimate of the CATE is just a dot product between $v(x_i)$ (transformed for the particular outcome) and the difference between the treatment and control embeddings. Formally:
$$\hat{\textrm{CATE}}_{jkt} (x_i) = v(x_i)' A_j (e^t_{k} - e^0_{k})$$

The fact that the CATE is fully linear is useful in real world systems where new experiments are constantly run. We will explore this idea in detail in section \ref{section-deployment}.

% !TEX root = main.tex
\section{Evaluation}

\subsection{How to Evaluate?}
Let $\hat{\text{ITE}}$ be the predicted ITE tensor from some model. Ideally, we would be able to calculate the loss between $\hat{\text{ITE}}$ and $\text{ITE}$, directly. This is often called the precision in estimating heterogeneous effects (PEHE) loss \citep{hill2011bayesian}. Formally, we write this as
$$ \text{PEHE} = \dfrac{1}{| ITE |} \sum_{ijk} (\hat{\text{ITE}}_{ijk}
- \text{ITE}_{ijk})^2.$$ 

Unfortunately, the true ITE is not observed (since any unit is ever only assigned to treatment or control in any experiment). Thus, there is an active area of research for proxy metrics which can be used instead of the true PEHE \citep{schuler2018comparison}.

We now describe two such candidates. Let $\mathcal{V}$ be a held out test set of unit, experiment, treatment, measure level outcomes with generic element $(ijkt) \in \mathcal{V}.$ Let $\hat{Y}$ be the full outcome tensor that is predicted by the model.

The first candidate is straightforward: it is the supervised learning objective for how well the model predicts the observed outcomes in the held out set. This is called the $\mu$-risk in the literature \citep{schuler2018comparison}. Formally this is given by $$\widehat{\mu\textrm{-risk}} = \frac{1}{|\mathcal{V}|} \sum_{(ijkt) \in \mathcal{V}} (\hat{y}_{ijkt} - y_{ijkt})^2$$

While $\mu$-risk is often used in practice, there are certainly situations where improvements in $\mu$-risk does not imply improvements in PEHE (for example, the simple case where base outcomes are predicatble but the treatment effect is constant). 

The second metric, called $\tau$-risk, is an unbiased proxy \citep{nie2020quasioracle, schuler2018comparison} for the ITE. It is defined as follows:

$$
\widehat{\tau\textrm{-risk}} = \frac{1}{|\mathcal{V}|} \sum_{x_i, y_i, t_i \in \mathcal{V}} ((y_i - \check{m}(x_i))-(t_i - \check{p}(x_i))\hat{\tau}(x_i))^2,
$$

\noindent where $\hat{\tau}(x_i)$ is the model estimate of $\tau$. 

Here $\check{m}$ is an estimate of $E[Y|X]$, and $\check{p}$ is the estimate of the propensity score both trained using data in the validation set. In our case both $\check{m}$ and $\check{p}$ are trained using linear models. In expectation $\tau$-risk is minimized when $\hat{\tau}$ is the true CATE function. However, for a full derivation we point the interested reader to \citet{nie2020quasioracle} and \citet{schuler2018comparison}.

Both $\tau$-risk and $\mu$-risk are proxy metrics for the unobservable ITE. Some work argues that $\tau$-risk is a better proxy than $\mu$-risk for evaluating heterogeneous treatment effect models, though the debate is not settled \citep{nie2020quasioracle, schuler2018comparison}. 

In the following sections, we always report the $\mu$-risk and report PEHE when ITEs are observed (in synthetic and semi-synthetic data) and the $\tau$-risk when PEHE is not observed.

\subsection{Baseline - Independent T-learners}

\begin{algorithm}[]
\caption{Independent T-learners}\label{algorithm:indt}
\begin{algorithmic}
\State $ML$ is your favorite machine learning model
\State $X \gets$ features for training set
\State $Y \gets$ outcomes from training set
\ForEach {experiment, $k$}
\ForEach {metric, $j$}
\State $X^1_{j,k} \gets$ features for units in treatment
\State $X^0_{j,k} \gets$ features for units in control
\State $Y^1_{j,k} \gets$ outcomes for units in treatment
\State $Y^0_{j,k} \gets$ outcomes for units in control
\State Learn $\hat{\mu}^1_{j,k} \gets ML(Y^1_{j,k} \sim X^1_{j,k})$ 
\State Learn $\hat{\mu}^0_{j,k} \gets ML(Y^0_{j,k} \sim X^0_{j,k})$
\State $\hat{\text{CATE}}_{j,k}(X) \gets \hat{\mu}^1_{j,k} (X) - \hat{\mu}^0_{j,k} (X)$
\EndFor
\EndFor
\end{algorithmic}
\end{algorithm}

As the baseline, we use the standard \textit{independent T-learner} approach. Algorithm \ref{algorithm:indt} provides pseudocode for training independent T-learners for each metric-treatment combination.

\subsection{Experiment 1: Synthetic Data}

We evaluate the LR-learner on synthetic experiment data with feature and outcomes generated as shown in algorithm \ref{algorithm:synth-data}.

\begin{algorithm}[]
\caption{Generating synthetic data}\label{algorithm:synth-data}
\begin{algorithmic}
\State $n \gets$ num samples per exp. condition
\State $d \gets $ latent feature dimension
\State $m \gets $ observed feature dimension
\ForEach {metric, $j$}
    \State $A_j \gets \texttt{randn}(d, d)$
\EndFor
\newline
\ForEach {experiment, $k$}
    \State $e^1_{k} \gets $ \texttt{l2\_norm}(\texttt{randn}(d))
    \State $e^0_{k} \gets $ \texttt{l2\_norm}(\texttt{randn}(d))
    \State $v \gets $ \texttt{fro\_norm}(\texttt{randn}(n, d))
    \ForEach {metric, $j$}
        \State $Y^1_{j,k} \gets v' A_j e^1_{k}$
        \State $Y^0_{j,k} \gets v' A_j e^0_{k}$
        \State $ITE_{j,k} = Y^1_{j,k} - Y^0_{j,k}$
    \EndFor
    \State $L \gets \texttt{randn}(d, m)$
    \State $X_k \gets v \cdot L$
    \State $T_k \sim \text{Bern}_n(0.5)$
\EndFor
\end{algorithmic}
\end{algorithm}

Using algorithm \ref{algorithm:synth-data}, we generate 50 synthethic data experiments with $m=128$ and 5 outcome metrics. Notice that our synthetic data assumes each unit is only a part of a single experiment, but this is not necessary for training the LR-learner.

For this data set we use an LR-learner without a non-linearity in the neural network with hidden dimension $d=32$. We also use linear regressions for our independent T-learner models. 

Since the data generating process (DGP) is linear by construction, both the independent T-learner and the LR-learner are able to learn the true DGP. The only difference between the two is that the LR-learner pools data across experiments and outcome measures.

In our analysis, we train both models on datasets of various sizes of per experiment data ($n=$ 25, 50, 100, 250, 500, 1000, 2500, 500, 10000, 50000 per treatment condition). We average over 5 independent replications.

\begin{table}[]
\centering
\resizebox{0.49\textwidth}{!}{%
\begin{tabular}{l|l|l|l|l|}
\cline{2-5}
 & \multicolumn{2}{c|}{$\textrm{PEHE}$} & \multicolumn{2}{c|}{$\mu\textrm{-risk}$} \\ \hline
\multicolumn{1}{|l|}{n} & \multicolumn{1}{c|}{\begin{tabular}[c]{@{}c@{}}Ind.\\ T-learners\end{tabular}} & \multicolumn{1}{c|}{LR-learner} & \multicolumn{1}{c|}{\begin{tabular}[c]{@{}c@{}}Ind.\\ T-learners\end{tabular}} & \multicolumn{1}{c|}{LR-learner} \\ \hline
\multicolumn{1}{|l|}{25} & 0.26792 & \textbf{0.02014} & 0.13407 & \textbf{0.01008} \\ \hline
\multicolumn{1}{|l|}{50} & 0.07956 & \textbf{0.02004} & 0.03962 & \textbf{0.01003} \\ \hline
\multicolumn{1}{|l|}{100} & 0.03292 & \textbf{0.02001} & 0.01651 & \textbf{0.01001} \\ \hline
\multicolumn{1}{|l|}{250} & 0.02307 & \textbf{0.02001} & 0.01154 & \textbf{0.01001} \\ \hline
\multicolumn{1}{|l|}{500} & 0.02144 & \textbf{0.02000} & 0.01072 & \textbf{0.01001} \\ \hline
\multicolumn{1}{|l|}{1000} & 0.02069 & \textbf{0.02000} & 0.01035 & \textbf{0.01001} \\ \hline
\multicolumn{1}{|l|}{2500} & 0.02028 & \textbf{0.02000} & 0.01014 & \textbf{0.01001} \\ \hline
\multicolumn{1}{|l|}{5000} & 0.02014 & \textbf{0.02000} & 0.01007 & \textbf{0.01001} \\ \hline
\multicolumn{1}{|l|}{10000} & 0.02007 & \textbf{0.02000} & 0.01004 & \textbf{0.01001} \\ \hline
\multicolumn{1}{|l|}{50000} & 0.02001 & \textbf{0.02000} & 0.01001 & \textbf{0.01000} \\ \hline
\end{tabular}%
}
\caption{\label{tab:risk-error-table-synth} \small The $\textrm{PEHE}$ and $\mu\textrm{-risk}$ averaged over all 50 synthethic data experiments and 5 outcome metrics evaluated on a validation set. For each value of $n$, data is drawn 5 times without replacement.}
\end{table}

Table \ref{tab:risk-error-table-synth} shows that the LR-learner performs better than the Independent T-learners on PEHE as well as $\mu\textrm{-risk}$ metrics, with $200\times$ less data. 

\subsection{Experiment 2: Semi-Synthetic Data}
We now consider evaluating the LR-learner on a semi-synthetic dataset. A major issue with real datasets is that we never observe the true ITEs as each unit is either in treatment or control. Thus, the true PEHE error is never observable. 

For this reason, using semi-synthetic data has become a popular methodology in theoretical causal inference \citep{dorie2019automated}. Semi-synthetic datasets have features that are generated from real data but have potential outcomes that are constructed via a known generated mapping,

To generate a dataset with multiple potential outcomes, we use a technique employed in the multi-armed bandit literature to transform a multi-class classification problem into a multi-outcome contextual bandit problem \citep{dudik2014doubly,agarwal2014taming}.

We start with an original multi-class dataset, and train a classifier to predict the classes. Then, for each sample $i$, we use the predicted class $j$ logit as the potential outcome $y_{ij}$. For the features for the semi-synthetic causal inference task, we can use the original features or some transformation thereof. For example, if the original classifier is a multi-layer neural network, we can use activations from an intermediate layer.

We use CIFAR100 as our base dataset \citep{krizhevsky2009learning}. For our base classifier, we use the simple neural network setup used in the PyTorch computer vision tutorial\footnote{We follow the setup here \href{https://pytorch.org/tutorials/beginner/blitz/cifar10_tutorial.html} except we use CIFAR100 instead of CIFAR10 and use 400 hidden units for the intermediate linear layers. Otherwise, the code we use is identical.}.

\begin{figure}[ht]
\centering
\includegraphics[width=0.4\textwidth]{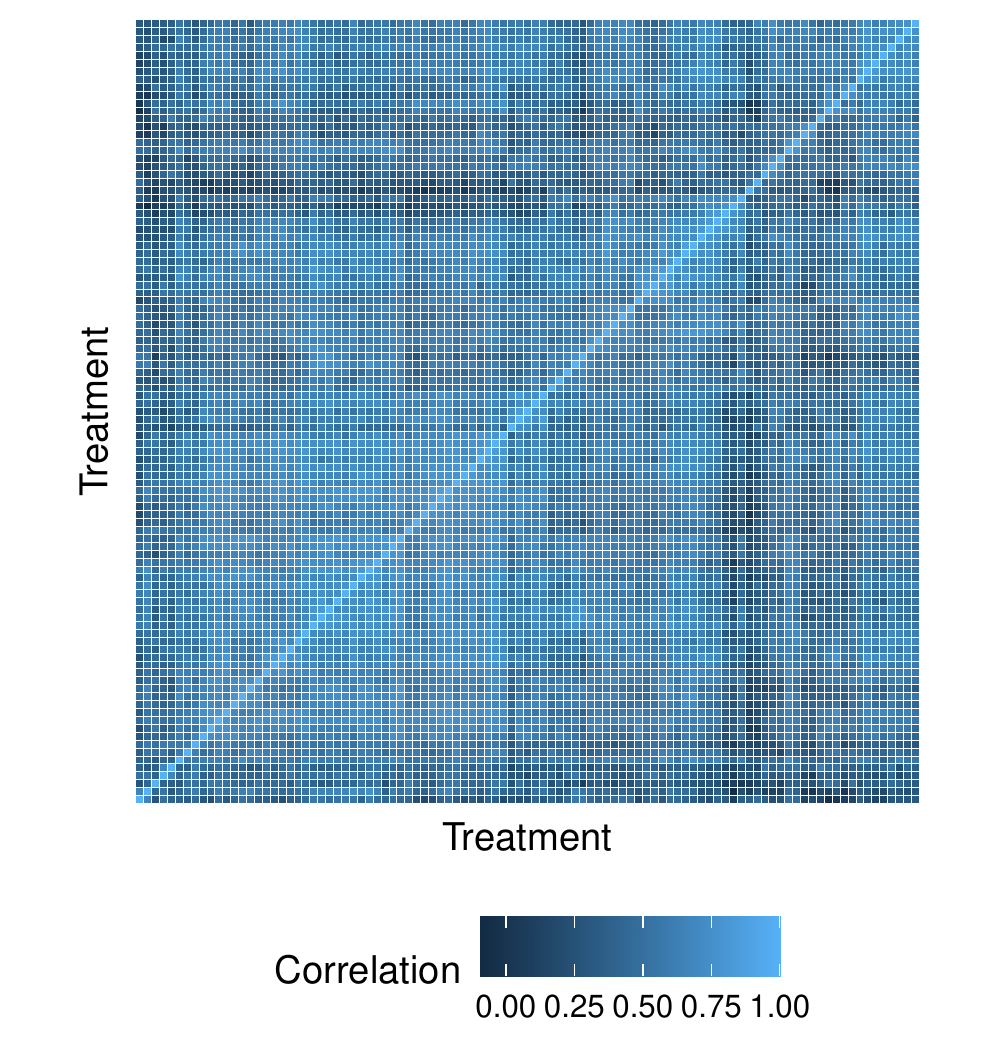} \\
\includegraphics[width=0.4\textwidth]{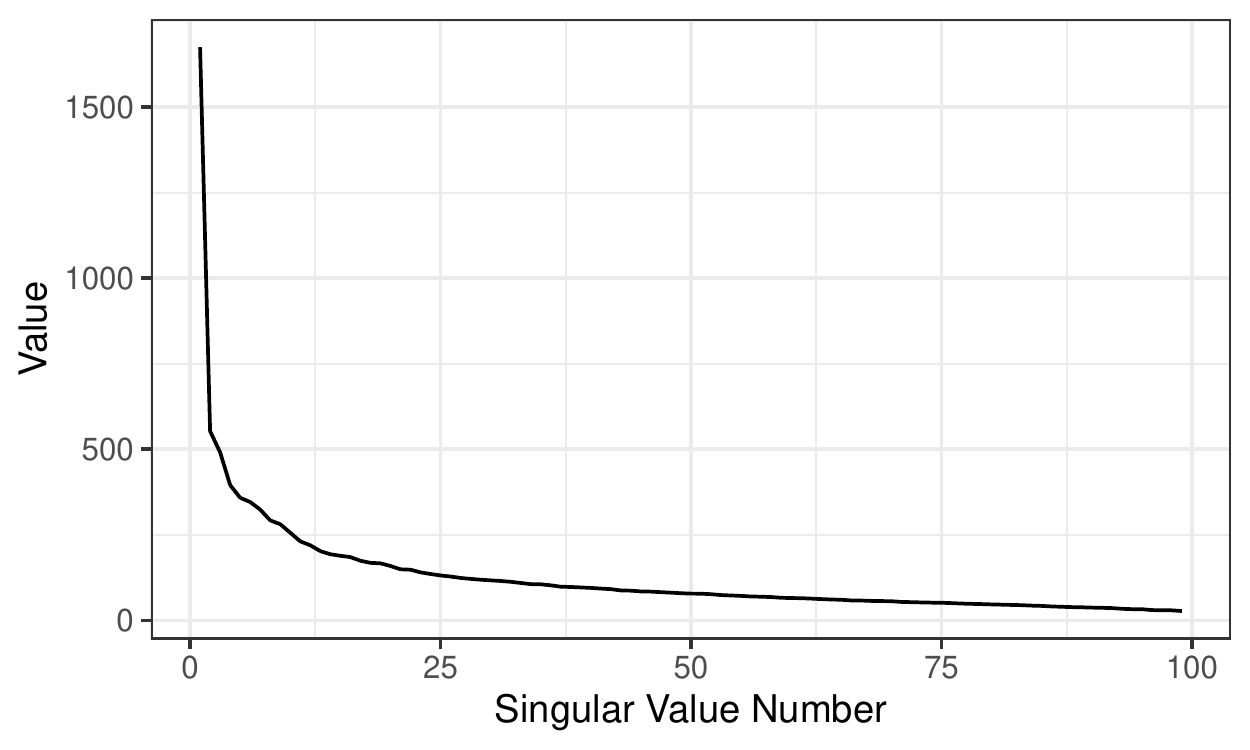}
\caption{The correlation matrix between the ITEs for each of the 99 treatments with hierchical clustering done on row/column ordering to emphasize the pattern as well as the singular value plot for the ITE matrix. We see that the ITE matrix induced by the CIFAR100 dataset is effectively low rank.}\label{cifar100_figure}
\end{figure}

For features, we use the $2^{nd}$ to last linear layer. For the semi-synthetic per-unit potential outcomes, we use the logit outputs for each class. This means that the real function, mapping features to logits, is a non-linear function.

Using this method on the CIFAR100 dataset gives us a semi-synthetic dataset with 400 features and 100 outcomes with 50000 units in our training set and 10000 units in our test set. We treat the first outcome as the control experiment and the other 99 experiments as the treatments. 

Figure \ref{cifar100_figure} shows the correlation matrix between the ITEs for each of the 99 treatments as well as the singular value plot for the ITE matrix. We see that the ITE matrix is effectively quite low rank. 

We randomly allocate 10 percent of the units to each experiment, so that on average each unit is a part of 10 experiments. Note that this semi-synthetic dataset only has one outcome metric.

We train a LR-learner model with a 2-layer neural network with ReLU nonlinearity and hidden dimension 128. As baseline, we train an Independent T-learner model using both a simple linear regression and XGBoost \citep{chen2016xgboost}. We tested other popular model classes (e.g. neural networks) for the independent T-learners but found that XGBoost performed best and required the least hyperparameter tuning to reach that level of performance. 

\begin{table}[]
\centering
\resizebox{0.3\textwidth}{!}{%
\begin{tabular}{l|l|l|}
\cline{2-3}
                                   & PEHE            & $\mu\textrm{-risk}$ \\ \hline
\multicolumn{1}{|l|}{LR-learner}    & \textbf{0.5277} & \textbf{0.2560}     \\ \hline
\multicolumn{1}{|l|}{XGBoost}      & 0.5787          & 0.2696              \\ \hline
\multicolumn{1}{|l|}{Linear Model} & 0.6118          & 0.2897              \\ \hline
\end{tabular}
}
\caption{\label{tab:risk-error-table-semi-synth} \small The $\textrm{PEHE}$ and $\mu\textrm{-risk}$ for the LR-learner model and XGBoost and linear model independent T-learner baselines. Values are averaged over all 100 semi-synthethic data experiments on a single outcome metrics evaluated on the test set.}
\end{table}

Table \ref{tab:risk-error-table-semi-synth} shows that the LR-learner performs better than the Independent T-learners (for both XGBoost and linear model baselines) on PEHE as well as $\mu\textrm{-risk}$ metrics. 
\subsection{Experiment 3: Large Scale Real Data}
We now turn to evaluating the LR-learner on a real dataset. We analyze a set of 53 randomized experiments run at Facebook during a 40 day period. These experiments are from various applications and are largely unrelated. In our sampled dataset, for each experiment we study $50,000$ deidentified users per treatment and control group in our training set, with an additional $10,000$ deidentified users in each group reserved for each of the test and validation sets. 

We consider 5 main outcomes, or key performance indicators (KPIs), core to Facebook. These experiments are not necessarily designed to move these outcome metrics. Each user may be in multiple experiments, but vast majority of users are in only a single experiment. 

As our $X_i$ we use a set of 142 features that include user embeddings trained for Facebook recommender systems, pre-treatment values of the KPIs, and user demographics. Users' pre-treatment features are taken before the 40 day period, and all post-treatment outcomes are recorded at the end of the 40 day period.

We define $v(x_i)$ as the output of a two-layered fully-connected neural network. The first layer takes in the set of 142 features as input and outputs a dimension $d$ representation, using a ReLU activation. The second layer outputs a dimension $d$ user embedding. The LR-learner is then trained over 250 epochs of the training data. 

This model is hyperparameter tuned on learning rate, weight decay and the model hidden dimension, $d$. We select the model with the lowest average risk across all 5 outcome measures computed on a validation dataset. When optimized for $\tau\textrm{-risk}$, we use a learning rate of 5e-4, weight decay of 5e-3, and $d=32$, and when optimized for $\mu\textrm{-risk}$, we use a learning rate of 1e-4, weight decay of 1e-4, and $d=64$.

We compare our LR-learner to a baseline of individual T-learners trained using XGBoost \citep{chen2016xgboost}. As above, we tested other popular model classes (e.g. neural networks) for the independent T-learners but found that XGBoost performed best and required the least hyper parameter tuning to reach that level of performance. 

The T-learners are trained on the same 142 feature set and are tuned with respect to the $\tau\textrm{-risk}$ or $\mu\textrm{-risk}$ on a held out datatest. Unlike the LR-learner where we use a single set of parameters with all experimental data pooled, we give the T-learner baseline an advantage by tuning \textit{each model} for a given experiment and metric separately on the following parameters: learning rate, learning rate decay, max tree depth, min child weight, gamma regularization, and fraction of features/observations subsampled while training each tree. 

\subsection{Results}
\begin{figure}
\centering
\includegraphics[width=0.49\textwidth]{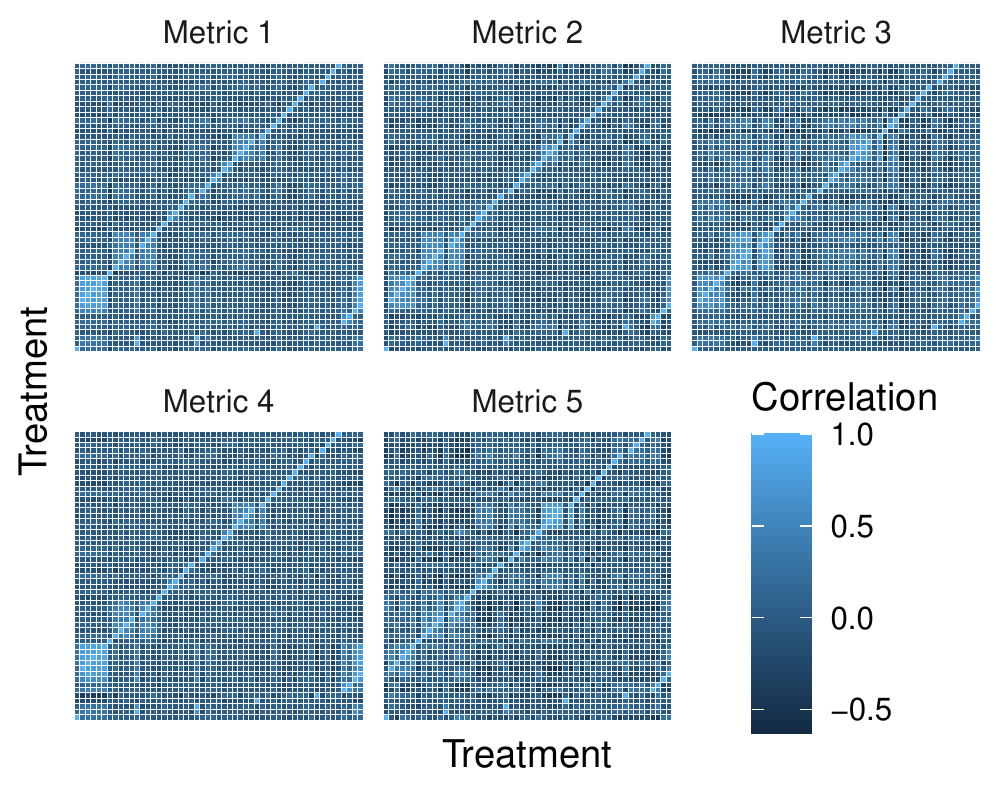} \\
\includegraphics[width=0.49\textwidth]{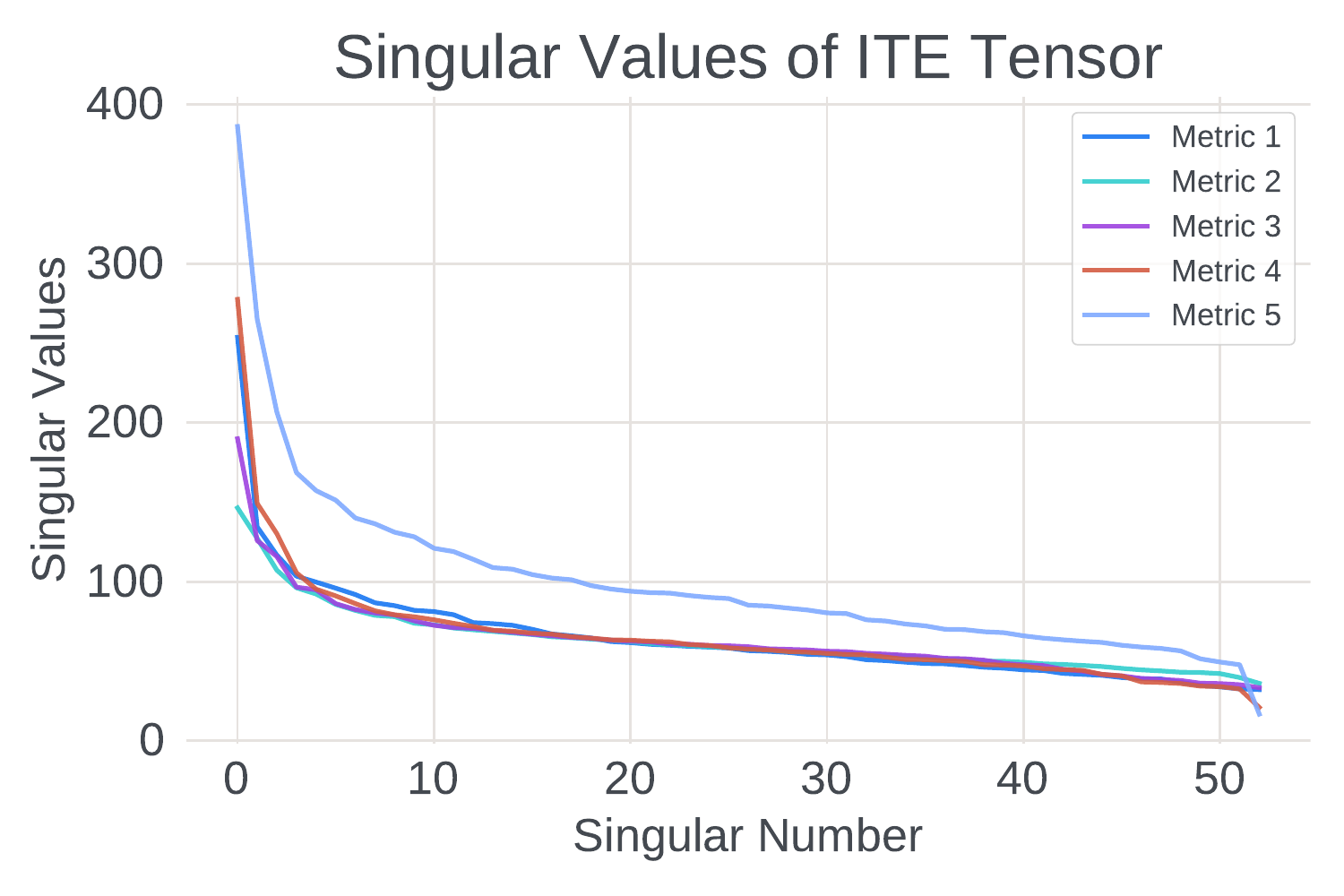}
\caption{\small The correlations between treatments of independently estimated ITE matrices for each of the $5$ outcome metrics and singular values of each ITE matrix suggest that the real ITE tensor is low rank.}\label{metric_figure}
\end{figure}

The key assumption that makes our procedure useful is that there is information that can be gained about CATEs across metrics and across experiments; i.e., that the underlying ITE tensor is low-rank. 

In our synthetic and semi-synthetic data, we could observe the ITEs/CATEs so we could verify this assumption in practice. In the real data, ground truth is unobserved, so we can assess the validity of the assumption by looking at the estimated ITE tensor from the independent T-learner procedure optimized for $\mu\textrm{-risk}$.

We take each slice of this tensor for each metric. Let $M^j$ denote such a slice. This is a matrix where a row is a unit and a column is an experiment and an entry is the unit's estimated ITE on that metric. 

Each column is estimated via an independent T-learner. Therefore, any correlation found in the column space is highly suggestive of an underlying correlation in true ITEs. 

We evaluate the rank of each $M^j$. The plot of each matrix's singular values (Figure \ref{metric_figure}) indeed suggests that the matrices appear to be effectively low rank. To quantify how low rank these matrices are, we use the bi-cross-validation procedure \citep{bcv} to estimate the effective rank of each matrix. Using Wold-style 5-fold cross-validation, we find that the effective rank for each outcome measure is $3, 5, 3, 4, 3$, respectively, confirming that all $M^j$ are indeed low-rank relative to the number of experiments.

Having established that the ITE tensor is low rank, we ultimately wish to quantify how the low-rank structure improves CATE estimation for the LR-learner. 

In Tables \ref{tab:risk-error-table-optmse} and \ref{tab:risk-error-table-opttau}, we calculate the $\tau\textrm{-risk}$ and $\mu\textrm{-risk}$ averaged over all 53 experiments, for each metric, using models optimized for $\tau\textrm{-risk}$ and $\mu\textrm{-risk}$, respectively. We notice that the LR-learner model optimized for $\mu\textrm{-risk}$ achieve relatively similar values of $\tau\textrm{-risk}$, but has vastly improved $\mu\textrm{-risk}$, which suggests that $\mu\textrm{-risk}$ is the more preferrable metric to optimize for. In table \ref{tab:risk-error-table-optmse}, the LR-learner outperforms the independent T-learners for both evaluation metrics, for all but one outcome. 

Note that the difference in $\mu$-risks (and in prior experiments PEHE) has a straightforward interpretation (when outcomes are scaled to variance $1$). The $\mu$-risk is just the out of sample percent of variance explained. By contrast, $\tau$-risk does not have such a simple interpretation since it is an unbiased but noisy proxy for the ITE. So, it is harder to say when there is a `big' or `small' difference in $\tau$-risk between two models.

\begin{table}[]
\centering
\resizebox{0.49\textwidth}{!}{%
\begin{tabular}{c|c|c|c|c|}
\cline{2-5}
 & \multicolumn{2}{c|}{$\tau\textrm{-risk}$} & \multicolumn{2}{c|}{$\mu\textrm{-risk}$} \\ \cline{2-5} 
 & \begin{tabular}[c]{@{}c@{}}Ind.\\ T-learners\end{tabular} & LR-learner & \begin{tabular}[c]{@{}c@{}}Ind.\\ T-learners\end{tabular} & LR-learner \\ \hline
\multicolumn{1}{|c|}{Metric 1} & 0.3747 & \textbf{0.3741} & 0.3637 & \textbf{0.3535} \\ \hline
\multicolumn{1}{|c|}{Metric 2} & 0.5566 & \textbf{0.5553} & 0.5515 & \textbf{0.5243} \\ \hline
\multicolumn{1}{|c|}{Metric 3} & 0.3219 & \textbf{0.3207} & 0.3270 & \textbf{0.3000} \\ \hline
\multicolumn{1}{|c|}{Metric 4} & 0.3977 & \textbf{0.3971} & 0.3749 & \textbf{0.3617} \\ \hline
\multicolumn{1}{|c|}{Metric 5} & \textbf{0.5357} & 0.5393 & 0.5354 & \textbf{0.4881} \\ \hline
\end{tabular}%
}
\caption{\label{tab:risk-error-table-optmse} \small The $\tau\textrm{-risk}$ and $\mu\textrm{-risk}$ averaged over all 53 experiments for each metric evaluated on a validation set. Models are parameter-tuned for $\mu\textrm{-risk}$.} 
\end{table}

\begin{table}[]
% \small
\centering
\resizebox{0.49\textwidth}{!}{%
\begin{tabular}{c|c|c|c|c|}
\cline{2-5}
 & \multicolumn{2}{c|}{$\tau\textrm{-risk}$} & \multicolumn{2}{c|}{$\mu\textrm{-risk}$} \\ \cline{2-5} 
 & \begin{tabular}[c]{@{}c@{}}Ind.\\ T-learners\end{tabular} & LR-learner & \begin{tabular}[c]{@{}c@{}}Ind.\\ T-learners\end{tabular} & LR-learner \\ \hline
\multicolumn{1}{|c|}{Metric 1} & 0.3743 & \textbf{0.3741} & \textbf{0.3734} & 0.3792 \\ \hline
\multicolumn{1}{|c|}{Metric 2} & 0.5557 & \textbf{0.5552} & 0.5741 & \textbf{0.5524} \\ \hline
\multicolumn{1}{|c|}{Metric 3} & 0.3212 & \textbf{0.3205} & 0.3496 & \textbf{0.3237} \\ \hline
\multicolumn{1}{|c|}{Metric 4} & 0.3974 & \textbf{0.3970} & \textbf{0.3897} & 0.4009 \\ \hline
\multicolumn{1}{|c|}{Metric 5} & \textbf{0.5310} & 0.5328 & 0.5701 & \textbf{0.5205} \\ \hline
\end{tabular}%
}
\caption{\label{tab:risk-error-table-opttau} \small The $\tau\textrm{-risk}$ and $\mu\textrm{-risk}$ averaged over all 53 experiments for each metric evaluated on a validation set. Models are parameter-tuned for $\tau\textrm{-risk}$.}
\end{table}

\section{Real World Deployment}
\label{section-deployment}
So far, we have considered the case of a fixed dataset $\mathcal{D}$, and an analyst interested in heterogeneous effects in all or a subset of the experiments/outcomes. We can also consider a scenario where the analyst already has access to an LR-learner trained on a large set of existing experiments. When a new experiment is run, the analyst may want to estimate HTEs for the new experiment. Call this new data $\mathcal{D}_{\text{new}}$. For simplicity, consider the case where there is just a single outcome metric of interest, and potentially this outcome measure is not one that is considered in past experiments.

One potential solution is to add the new experiment to the data, call this combined dataset $\mathcal{D}_{c}$ and re-train the LR-learner on the complete dataset. However, when $\mathcal{D}$ is extremely large relative to $\mathcal{D}_{\text{new}}$, then the addition of the new data will not change the shared parameters across all experiments much. Thus, we can simply fix the parameters of the neural network, $v_{\mathcal{D}}(\cdot)$, to the ones of the model trained on the original data. In other words, we can find new experiment treatment/control vectors $e_{\text{new}}$ for the outcome in question by solving:

\begin{equation*}
    \min_{e_{\text{new}}} \sum_{(y_{it},x_i) \in \mathcal{D}_{\text{new}}} (y - v_{\mathcal{D}} (x_i)' e^t_{\text{new}})^2.
\end{equation*}

In other words, we simply construct an independent T-learner on the new experiment. However, instead of using the original feature space $x$ and possibly a complex model, we use the pre-trained LR-learner as a feature extractor and a linear regression as the underlying model. This reduces the complexity of the learning problem substantially.  

This is very similar to the idea of fine-tuning supervised (or unsupervised) models in other areas of machine learning such as vision or NLP where an embedding layer of a model trained on a large dataset is used as the feature space for a simpler model in a new dataset \citep{mikolov2013distributed,mahajan2018exploring}.

In practice, if an LR-learner were to be deployed in an organization, we would suggest the following process: Retrain the full model on a pool of experiment data intermittently. Save the model, and in particular, save the feature extractor $v(\cdot)$. In daily analyst practice, when estimating heterogeneous treatment effects for a new experiment, reuse the saved $v(\cdot)$ to train linear independent T-learners.

\section{Conclusion}

In this paper, we devise the LR-learner, an approach for simultaneous CATE estimation across multiple experiments and multiple outcomes. Our tensor factorization approach assumes a low rank structure in the ITE tensor across experiments and outcome metrics. We show that there is indeed a low-rank structure for CATEs across experiments and outcomes, which allows the LR-learner to be $200 \times$ more sample efficent compared to individual CATE models. We show using real data from Facebook that the LR-learner outperforms the Independent T-learner approach in both $\tau\textrm{-risk}$ and $\mu\textrm{-risk}$ evaluation metrics.

Our approach can be especially useful in domains such as internet companies where experiments are continuously being run, potentially across various unrelated applications, and outcome metrics are inexpensive to obtain. This is in contrast to many fields like medicine in which panel data is more common, and in which experiments may be even more related, but multiple outcome metrics can be difficult to obtain. One avenue for future work is to evaluate the LR-learner on panel data and compare it to other causal matrix completion methods \citep{agarwal2021causal, athey_causal_factorization}.

We have focused purely on the case of A/B tests, but another fruitful direction is to combine the LR-learner architecture with datasets that are either wholly or partially observational \citep{shalit2017estimating,lada2019observational}. The architecture naturally extends to estimating treatment assigment probability as well as outcomes, so a possible future direction is incorporating either X-learner \citep{kunzel2019metalearners} or doubly-robust \citep{kennedy2020optimal} methods to the LR-learner.

We establish the contribution of the LR-Learner as a method for estimating CATEs, and we hope that future work explores potential applications for the learned embeddings. The user, metric, and experiment embeddings have many potential downstream uses, such as finding or comparing  similar users or experiments. The learned embeddings may also inform the design of future experiments.

\bibliographystyle{ACM-Reference-Format}
\bibliography{refs}

\end{document}